\documentclass[11pt,a4paper]{article}
\usepackage[hyperref]{acl2021}
\usepackage{times}
\usepackage{latexsym}

\usepackage{microtype}

\usepackage{amsmath}
\usepackage{graphicx}
\usepackage{multirow}

\aclfinalcopy

\title{THINK: A Novel Conversation Model for Generating Grammatically Correct and Coherent Responses}

\author{Bin Sun$^1$, Shaoxiong Feng$^1$, Yiwei Li$^1$, Jiamou Liu$^2$, Kan Li$^{1*}$\\
  $^1$School of Computer Science \& Technology, Beijing Institute of Technology \\
  $^2$School of Computer Science, The University of Auckland \\
  \texttt{\{binsun,shaoxiongfeng,liyiwei,likan\}@bit.edu.cn} \\
  \texttt{jiamou.liu@auckland.ac.nz}}

\date{}

\begin{document}
\maketitle
\begin{abstract}
Many existing conversation models that are based on the encoder-decoder framework have focused on ways to make the encoder more complicated to enrich the context vectors so as to increase the diversity and informativeness of generated responses. 
However, these approaches  face two problems.
First, the decoder is too simple to effectively utilize the previously generated information and tends to generate duplicated and self-contradicting responses.
Second, the complex encoder tends to generate diverse but incoherent responses because the complex context vectors may deviate from the original semantics of context.
In this work, we proposed a conversation model named ``THINK'' (Teamwork generation Hover around Impressive Noticeable Keywords) to make the decoder more complicated and avoid generating duplicated and self-contradicting responses. The model simplifies the context vectors and increases the coherence of generated responses in a reasonable way. For this model, we propose \textit{Teamwork generation} framework and \textit{Semantics Extractor}. Compared with other baselines, both automatic and human evaluation showed the advantages of our model.
\end{abstract}

\section{Introduction}

Open-domain dialogue generation is a challenging task of natural language process (NLP). Early chat-bots tend to generate generic and dull responses. During the past few years, many neural conversation models have been proposed that leverage the power of the encoder-decoder architecture ~\citep{encoder-decoder-1,encoder-decoder-2}. These models generate diverse responses with reduced blandness and dullness through more complex context vectors ~\citep{attention-origin-bahdanau15,attention-origin-luong15,transformer-base-17,cmham-multi-head-attention,external-info-jiwei,external-info-Marjan,external-info-Bernd,vae-bowman16,cvae-zhao17}. However, these work still face problems.

\begin{table}[!t]
\renewcommand{\arraystretch}{1.3}
\centering
\begin{tabular}{rl}
\hline
 & \textbf{I’m afraid} I’m afraid I’m afraid.  \\
Duplicated & yes, \textbf{sir, sir, sir, sir, sir, sir, sir,} \\
 responses:& I never see it \textbf{either either either}.\\
\hline
 Self-contr-& \textbf{I love} this movie but \textbf{I don’t love}\\
 -adicting& this movie.  \\
 responses:& Hi, \textbf{I’m leonard. I’m melinda.} \\
\hline
\end{tabular}
\caption{Duplicated and self-contradicting responses generated by conversation models with simple decoders.}
\label{tab:samples}
\end{table}

First, despite the added complexity to the context vectors, existing encoder-decoder approaches still employ relatively simpler decoders, \emph{e.g.}, long-short term memory (LSTM) ~\cite{lstm97}, gated recurrent unit (GRU) ~\cite{gru2014}, and transformer decoder. Many studies and empirical evidences ~\citep{attention-origin-bahdanau15,transformer-base-17,cmham-multi-head-attention,cvae-zhao17} have shown that these decoders tend to generate duplicated and self-contradicting responses (see Table~\ref{tab:samples}).
\begin{figure}[ht]
\centering
\includegraphics[scale=0.4]{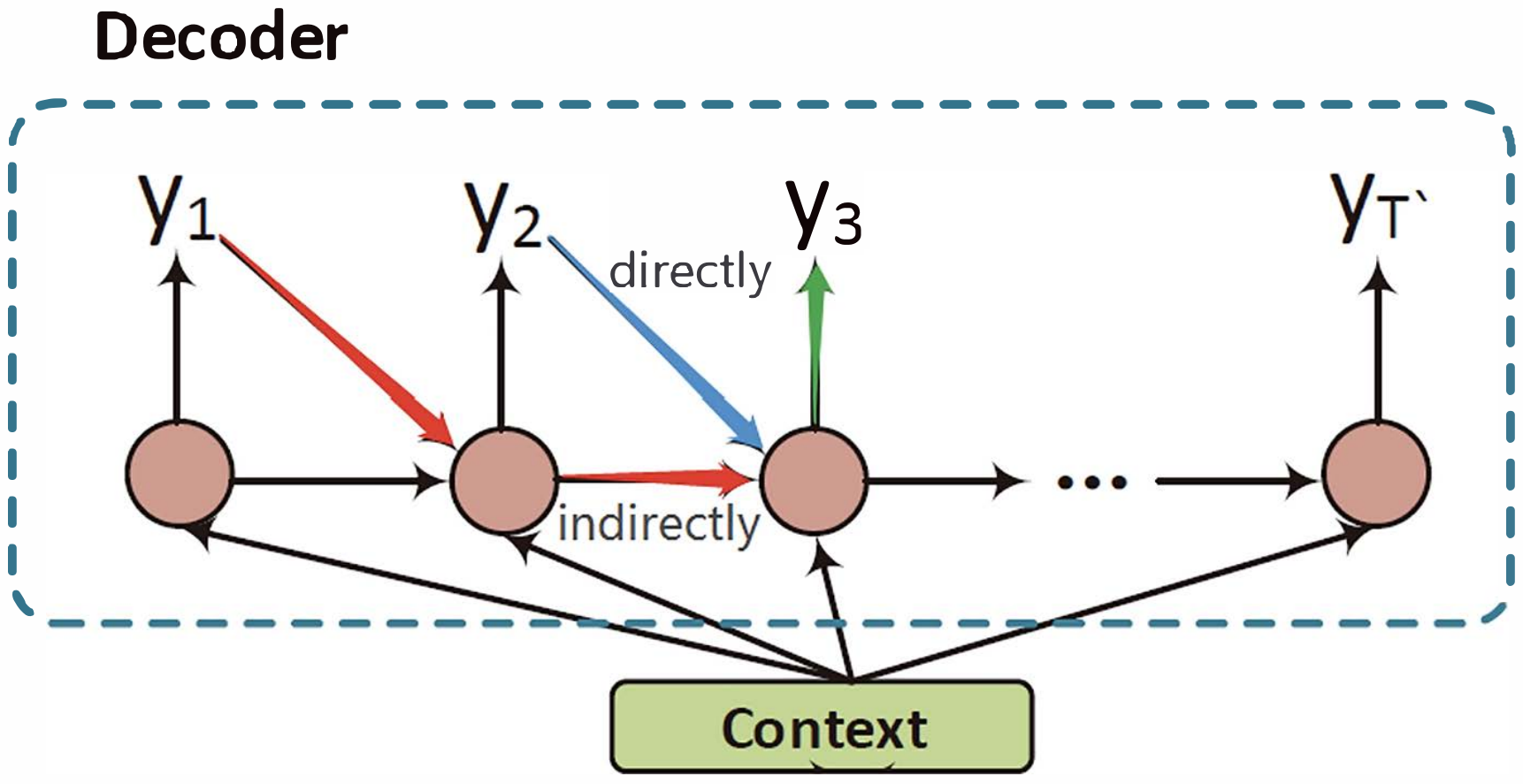}
\caption{An illustration of the decoding process of Encoder-Decoder framework \cite{reviewChen2017}}
\label{fig-ed-architecture}
\end{figure}
As shown in Figure~\ref{fig-ed-architecture}, in the decoding process of the encoder-decoder framework, the previously generated tokens $y_1, y_2,\ldots, y_{i-2}$ are indirectly used to generate $y_i$. For LSTM or GRU decoder, the indirect input information has to pass through the ``gate'', which makes it easy to be changed or be forgot. For the transformer decoder, the indirect input information will be reconstructed through the self-attention mechanism ~\citep{transformer-base-17}. Since the attention mechanism selects information generally through the inner product, some important information will be ignored during the computation.

Second, high complexity of the context vectors often negatively impact the coherence of the generated responses. Past encoder-decoder approaches often focus on making the context vectors more complicated with external messages \citep{external-info-jiwei,external-info-Bernd,external-info-Marjan} or latent variables \citep{vae-bowman16,cvae-zhao17,spacefusion-gaoxiang19}. However, the single-turn dialogue datasets hardly contain an external message, resulting in the latent variables easily producing incoherent responses. For instance, when given the post \textit{Everything about this movie is awesome!} as context, the CVAE model generated \textit{Caves would never say yes, but I’d love to know.} \citep{spacefusion-gaoxiang19}.

Thus, the simplicity of the decoder coupled with the high complexity of the context vectors often lead to grammatically-incorrect and incoherent generate responses. To address this problem, we propose  \textit{Teamwork Generation} framework (\textit{TG}), and \textit{Semantics Extractor} (\textit{SE}). \textit{TG}  makes the decoding process more complex by directly and explicitly utilizing the previous generated tokens during the generating process. This would effectively mitigate the problem of the generated responses being duplicated and self-contradicting. Inspired by the divide-and-conquer strategy and Map-Reduce framework, \textit{TG} divides the task of generating one response into the task of generating $N$ tokens (see Section ~\ref{sect:TG_framework} for details). 

\begin{figure}[ht]
\centering
\includegraphics[scale=1.0]{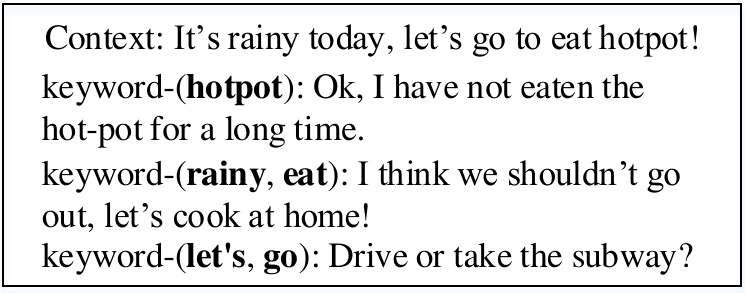}
\caption{Conversation samples that various responses consider different keyword-combination}
\label{fig-example}
\end{figure}

\textit{SE} is used to simplify the context vectors. Examining the phenomena as shown in Figure \ref{fig-example} and sentence classification task \citep{TextCNN-kim-2014,DCNN-Kalchbrenner-2014,DilatedConv-2016-ICLR}, we find that keywords are more important than other word for understanding sentence semantics. Therefore, we propose \textit{SE} to extract the possible keyword-combinations that are semantically-linked with possible responses (see Section ~\ref{sect:SE} for details), which could effectively increase the coherence between responses and contexts.

In this work, we focus on the {\em single-turn} dialogue generation task, \emph{i.e.}, giving a post (context) and generating a response, and proposed a novel conversation model based on the proposed \textit{Teamwork generation} framework and \textit{Semantics Extractor}. We named it \textit{Teamwork generation Hover around Impressive Noticeable Keywords} (\textit{THINK}). The model is significantly different from existing conversation models. Our contributions are summarized as follows: (1) We propose a new End-to-End framework (named \textit{Teamwork generation}) to ensure that all previous generated tokens can be directly utilized during generating process, which can effectively address the duplicate and self-contradicting responses.
(2) We propose a \textit{Semantics Extractor} that is based on the deformable convolution idea to extract semantic features, which is helpful to generate coherent responses. This is the first study to our knowledge of the utilization of deformable convolution idea in dialog generation task.
(3) We perform detailed experiments to verify the ability of \textit{THINK} on generating grammatically-correct and coherent responses.

\begin{figure*}[ht]
\centering
\includegraphics[scale=0.6]{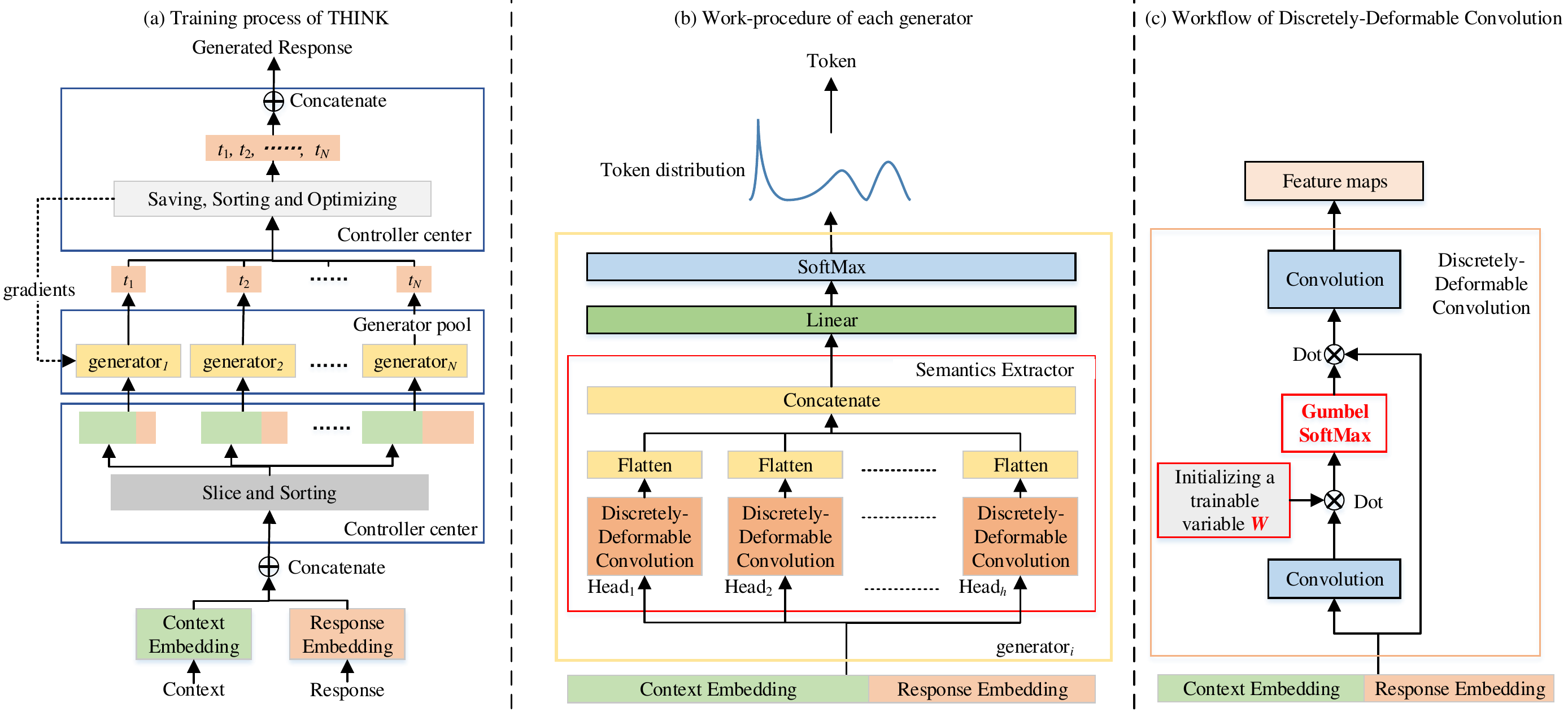}
\caption{An illustration of the \textit{THINK} model. Figure~\ref{fig-think}(a) shows the training process of the Teamwork generation framework; Figure~\ref{fig-think}(b) is the work-procedure of each generator; Figure~\ref{fig-think}(c) is the workflow of the Discretely-Deformable Convolution.}
\label{fig-think}
\end{figure*}

\section{Model}

\subsection{Overview}
\label{sect:def-conv-overview}

Figure~\ref{fig-think} illustrates the framework of the \textit{THINK} model.
 \textit{TG}  works by saving all previous generated tokens and directly utilizing them, which could effectively address the problem with responses being duplicated and self-contradicting. \textit{SE} extracts impressive and noticeable keywords instead of complex context vectors, which is helpful to increase coherence of the generated responses.

\subsection{Teamwork generation framework}
\label{sect:TG_framework}

A dialogue generation model using the encoder-decoder framework generates duplicated and self-contradicting responses as the decoder forgets the previously generated tokens. Therefore, to mitigate this problem, we proposed the \textit{TG} framework.

The \textit{TG} framework consists of two components: a \textit{Generator pool} and a \textit{Control center}, which are shown in Figure \ref{fig-think}(a). The \textit{Generator pool} contains $N$ generators and force each generator to only generate one token. The \textit{Control center} is responsible for managing the inputs and outputs of all generators. First, the \textit{Control center} gives each generator a sequence number and requires the $i$th generator to generate the $i$th token of a response. In addition, the \textit{Control center} also saves all generated tokens, and put the 1st, 2nd, $\ldots$, $(i-1)$th generated tokens as input of the $i$th generator. The steps of  \textit{TG} is shown in \eqref{eq:1}.
\begin{align}
\label{eq:1}
    &response = Concatenate(t_1, t_2,...,t_{N})\\
    \nonumber &t_i = idx2word(\arg\max(generator_i(\mathbf{X}_i))) \\
    \nonumber &\mathbf{X}_i = Embed(Concatenate (C,t_1, t_2,...,t_{i-1})),
\end{align}
where $response$ is the generated response of the original context $C$, $t_i$ denotes the $i$th token of the response, $generator_i$ denotes the $i$th generator that is responsible for generating $t_i$. The function $idx2word$ is used to convert the probability into token, and $\mathbf{X}_i$ represents the embedding result of the $i$th context. When $C$ is input, the $generator_1$ will generate $t_1$. \textit{Control center} then combines $t_1$ with $C$ as the $2$nd context and inputs it to $generator_2$ to obtain  $t_2$. This is repeated until the last generator produces $t_N$ after which the controller concatenates all tokens as a response.

\subsection{Generator}
\label{sect:generator}

We employ \textit{SE} and a Multilayer Perceptron (MLP) as the generator, which is shown in Figure~\ref{fig-think}(b). \textit{SE} first extracts the semantics features $\mathbf{f_{se}}$ from the input sentence embedding $\mathbf{X}$, and then the MLP utilizes $\mathbf{f_{se}}$ to compute the probability of output tokens $p_t$, which is shown in \eqref{eq:2}.
\begin{equation}
\label{eq:2}
    p_t = MLP(\mathbf{f_{se}}), where\ \mathbf{f_{se}}=SE(\mathbf{X})
\end{equation}

\subsection{Semantics extractor}
\label{sect:SE}

Convolutional neural network (CNN) is popular for extracting the semantic features in sentence classification tasks \citep{TextCNN-kim-2014,DCNN-Kalchbrenner-2014,DilatedConv-2016-ICLR,DepthwiseConv-Kaiser-2018}. Therefore, we propose \textit{Semantics Extractor}  (\textit{SE}) which extracts relevant semantic information from context sentence using CNN. Due to inability to capture long-distance dependent features,  the standard CNN   will cause certain problems when it is directly applied to  NLP tasks. Existing methods mainly apply two strategies to mitigate this problem with CNN: the first is to increase the depth of the CNN model, and the other is to change the convolution kernel (\emph{e.g.}, Dilated Convolution).

To extract long-distance dependent features, we propose \textit{multi-head discretely-deformable convolution} as \textit{SE}, which is inspired from the deformable convolution \citep{deformable-conv-daijifeng-17}, gumbel softmax \citep{gumbel-softmax-17} and multi-head strategy \citep{cmham-multi-head-attention}. The workflow is shown as \eqref{eq:3}.
\begin{align}
\label{eq:3}
    &\nonumber \mathbf{f_{se}} = \mathrm{Concatenate}(\mathbf{f}_{final}^1, \mathbf{f}_{final}^2,...,\mathbf{f}_{final}^{head}) \\
    &\nonumber \mathbf{f}_{final}^i = Flatten(DiscretelyDC_i(\mathbf{X}_i,\mathbf{K}{final})) \\
    &and\ \mathbf{K}_{final}=(k, m, 1, p),
\end{align}
where $DiscretelyDC$ denotes the proposed \textit{discretely-deformable convolution} and $\mathbf{K}_{final}$ denotes its parameters. Two hyper-parameters $head$ and $p$ for adjusting the effect of the results: $head$ is the number of $DiscretelyDC$, and $p$ is the channel of the final convolution operation.

\subsubsection{Discretely-deformable convolution}
\label{sect:ddc}

The original deformable convolution trains an offset to make the convolution kernel shift to the middle position of the sentence embedding matrix, and it introduces too much unreasonable features and semantic errors for dialogue generation task. Therefore we modify the original definition and propose \textit{discretely-deformable convolution} (\textit{DiscretelyDC}) which is more suitable for handling sentences.  \textit{DiscretelyDC} trains a translation matrix, which makes sure that the deformable process only focuses on the relationship between words rather than the dimensions of the vector, to replace the offsets and linear interpolation. Figure~\ref{fig-think}(c) shows the implementation of \textit{DiscretelyDC}.

For a given sentence embedding matrix $\mathbf{X}_{n \times m}$ (where $n$ represents the actual number of tokens and $m$ represents the dimension of word embedding), we first perform a convolution operation with $\mathbf{K}$ ($k$, $m$, $1$, $n$), where $k$ is an adjustable parameter indicating the receptive field size.
The $stride$ size of convolution is set as 1.
\begin{align}
    \mathbf{f}=Conv(\mathbf{X},\mathbf{K},stride)
\end{align}

We then transpose $\mathbf{f}$ to form a matrix $\mathbf{M}_{f}^T$, and construct a trainable variable $\mathbf{W}$.
\begin{align}
    \mathbf{P}_{n \times n} = SoftMax(\mathbf{M}_f^T * \mathbf{W})
\end{align}

A probability matrix $\mathbf{P}_{n \times n}$ is obtained and $p_{ij} \in \mathbf{P}_{n \times n}$ denotes the probability that the token at position $j$ will shift to position $i$. Then, we use \textit{gumbel softmax} to implement the discretization process.
\begin{align}
    \mathbf{P}^h_{n \times n} = discrete\_process(\mathbf{P}_{n \times n})
\end{align}

Since we focus on the extreme situations, it can be simple as $P^{h}_{n \times n}=stop\_graidnet(\mathbf{P}^{\arg\max}_{n \times n}-\mathbf{P}_{n \times n})+ \mathbf{P}_{n \times n}$, where the $\mathbf{P}^{\arg\max}_{n \times n}$ is the results after argmax process.
The elements of $\mathbf{P}^h_{n \times n}$ are 0 or 1.
Then, we get the new $X_{deform}$:
\begin{align}
    \mathbf{X}_{deform} = \mathbf{P}^h_{n \times n} * \mathbf{X}
\end{align}

Finally, after the last convolution operation with $\mathbf{K}_{final}=(k, m, 1, p)$, we get the final feature:
\begin{align}
    \mathbf{f}_{final}=Conv(\mathbf{X}_{deform}, \mathbf{K}_{final}, stride)
\end{align}

\subsection{Training step}
\label{sect:Training}

During training, we use the TeacherForcing \cite{teacherforcing89} method with the loss function shown in \eqref{eq:loss}:
\begin{align}
\label{eq:loss}
    &\nonumber loss = CrossEntroy(logits, R) \\
    &\nonumber logits = Concatenate(logit_1, logit_2,...,logit_{r\_len})\\
    &\nonumber logit_i = p_t^{gi} = generator_i(\mathbf{X}_i)\\
    &\nonumber \mathbf{X}_i = Slice(\mathbf{X}_{UE},i+c\_len-1) \\
    &\mathbf{X}_{UE} = Embed(Concatenate(C, R))
\end{align}
where $R$ is true response, $C$ is the context sentence, $c\_len$ and $r\_len$ means the length of $C$ and $R$ respectively. For improving the training speed and preventing overfitting, we employ L2 regularization and label smoothing ~\cite{label-smoothing-16}.

\section{Experiments}
\label{sect:experiments}

\subsection{Dataset}
\label{sect:dia-data}
We use two public dialogue datasets in our experiments and modify them as single-turn dialog data.
The first dataset, named DailyDialog ~\cite{dailydialog}, consists of dialogues that reflect human daily communication patterns and cover various topics about our daily life.
After processing, the training, validation and testing sets contain 76,052, 7,069 and 6,740 pairs, respectively.
The second dataset is extracted from raw movie scripts, named CornellMovie ~\cite{cornellmovie}, finally left 77,697 pairs for training set, 9,671 for validation set, and 9,650 for testing set.

\subsection{Evaluation metrics}
\label{sect:dia-met}

For auto-evaluation metrics, we choose the \textbf{distinct}-$n$ \citep{distinct-16} and \textbf{coherence} \citep{coherence}. However, these metrics are not perfect. Since  \textbf{distinct}-$n$ only focuses on the quantity of distinct $n$-grams and overlooks the quality of $n$-grams, even the random generated response gets a high \textbf{distinct}-$n$ value \citep{spacefusion-gaoxiang19}, which couldn't evaluate the duplicate and self-contradict responses. Meanwhile, the \textbf{coherence} only focus on the sentence embedding level, which is not comprehensive enough. Therefore, based on these metrics, we propose two better metrics: the \textbf{q\_phrase}-$n$ to evaluate the quality of responses on phrases level, and the \textbf{mix\_coh} to evaluate the coherence of responses. (see follow subsections for details)

For human evaluation, 200 randomly sampled contexts and their generated responses are given to three crowd workers, who are required to give the comparison results (\emph{e.g.} win, loss or tie) between our model and baselines based on quality.
The quality means that the response is syntactically correct and has a good correlation to its context.

\subsubsection{q\_pharse}
Different from \textbf{distinct}, we propose the \textbf{q\_phrase-$n$} which is calculated through \eqref{eq:q_phrase}.
\begin{align}
\label{eq:q_phrase}
    q\_phrase(n) = \frac{effective(ngrams)}{total\_occur(ngrams)}
\end{align}
The $effective(ngrams)$ is an intersection of $unique(ngrams)$ and $vocab(ngrams)$, where $unique(ngrams$ is a set of unique $n$-grams and $vocab(ngrams)$ is a set that represents the $n$-grams vocabulary extracted from the real data.
We suppose that $vocab(ngrams)$ hardly includes duplicate and self-contradict $n$-grams, so the $effective(ngrams)$ only contains $n$-grams without grammatical errors.
For example, the \textbf{distinct}-3 of (``\textit{I am fine}'', ``\textit{I are fine}'', ``\textit{are fine I}'', ``\textit{I are you}'', ``\textit{are are are}'') is 1, while \textbf{q\_phrase}-3 is 0.2.
Since only ``\textit{I am fine}'' is a grammatically correct phrase, the \textbf{q\_phrase}-3 is more reasonable.
\begin{table*}[!t]
\renewcommand{\arraystretch}{1.3}
\centering
\small
\begin{tabular}{lccclcccc}
\hline
\textbf{Model} & \textbf{q\_phrase-3} & \textbf{q\_phrase-4} & \textbf{q\_phrase-5} & & \textbf{avg(B)} & \textbf{avg(E)}& \textbf{coherence}& \textbf{mix\_coh} \\
\hline
Seq2Seq     & 0.1554 & 0.1538 & 0.1170 & & 0.2267 & 0.4744 & 0.4885 & 0.5781           \\
CVAE        & 0.2453 & 0.1513 & 0.0985 & & 0.2536 & 0.4734 & 0.5116 & 0.6082           \\
TransFM     & 0.2946 & 0.2349 & 0.1542 & & 0.2532 & 0.4533 & 0.4855 & 0.5888           \\
CMHAM       & 0.3053 & 0.2739 & 0.1996 & & \textbf{0.2684} & 0.4560 & 0.4917 & 0.6044  \\
\hline
THINK       & \textbf{0.3620} & \textbf{0.3326} & \textbf{0.2662} & & 0.2680 & \textbf{0.4757} & \textbf{0.5117} & \textbf{0.6205}   \\
\hline
Seq2Seq     & 0.1083  & 0.1055  & 0.0770 & & 0.2405 & 0.4599 & 0.4891 & 0.5854   \\
CVAE        & 0.1766  & 0.0744  & 0.0234 & & 0.2530 & 0.4503 & 0.4903 & 0.5913   \\
TransFM     & 0.2533  & 0.1953  & 0.1043 & & 0.2624 & 0.4494 & 0.4961 & 0.6006   \\
CMHAM       & 0.2245  & 0.1946  & \textbf{0.1293} & & 0.2657 & 0.4540 & 0.4966 & 0.6053   \\
\hline
THINK   & \textbf{0.2731}  & \textbf{0.2140}  & 0.1112  & & \textbf{0.2668} & \textbf{0.4652} & \textbf{0.5126} & \textbf{0.6175}  \\
\hline
\end{tabular}
\caption{Results of automatic evaluation metrics on DailyDialog dataset and CornellMovie dataset.}
\label{tab:auto-evluation-results}
\end{table*}

\subsubsection{mix\_coh}

The \textbf{mix\_coh} consists of three levels: word level, word embedding level, and sentence embedding level.
For word level, we choose BLEU and take the average value of BLEU-\{1,2,3\} as the \textbf{avg(B)}.
Besides, we employ the embedding-based metrics to reflect the correlation of word embedding level.
Using the average value of embedding-\{greedy, average, extrema\} as the \textbf{avg(E)}.
As for sentence embedding level, we adopt the \textbf{coherence}.
\begin{align}
\label{eq:mix-coh}
    &\nonumber mix\_coh_i = B\_score_i + E\_score_i + C\_score_i \\
    &\nonumber B\_score_i = avg(B_i)/\sum\nolimits_{i\in M} avg(B_i) \\
    &\nonumber E\_score_i = avg(E_i)/\sum\nolimits_{i\in M} avg(E_i) \\
    &C\_score_i = Coherence_i/\sum\nolimits_{i\in M} Coherence_i
\end{align}

In \eqref{eq:mix-coh}, $M$ is a set of models in this paper. $mix\_coh_i$ represents the \textbf{mix\_coh} of model$_i$.

\subsection{Training details}
\label{sect:dia-para}
We set the vocabulary size to 23,000 and 32,000 for DailyDialog and CornellMovie, respectively.
For a fair comparison among all models, we employ 256-dimensional word embeddings.
The numbers of hidden nodes are all set as 256, and batch sizes are set as 64.
The $c\_len$ and the $r\_len$ are all set as 25.
The $head$ is set as 6 and $p$ is set as 8.
Adam is utilized for optimization.
The $init\_lr$ is set to be 0.001, and $w\_step$ is set as 4000.
We run all models on a Titan RTX GPU card with Tensorflow.
We train all models in 100 epochs.
And we use greedy search to generate responses for evaluating.

\subsection{Baseline models}
\label{sect:dia-baselines}

We compare the proposed model with the following baseline models: Seq2Seq with attention \citep{attention-origin-bahdanau15}, CVAE \citep{cvae-zhao17}, Transformer (TransFM) \citep{transformer-base-17} and CMHAM \citep{cmham-multi-head-attention}.

\section{Results and Analysis}
\label{sect:result-dia-mestric}

\subsection{Results of conversation models}
\label{sect:result-dia-mestric}

\subsubsection{Auto-evaluation metrics}
\label{sect:result-dia-mestric}

Table \ref{tab:auto-evluation-results} shows the results of auto-evaluating metrics. We can see that on the \textbf{q\_pharse}-{3,4,5} metric, our \textit{THINK} almost achieved the best results on both two dataset. This result demonstrates that our \textit{THINK} generates the most diverse and meaningful {3,4,5}-grams (phrases).

Meanwhile, the \textit{THINK} has the best performance on both \textbf{mix\_coh} and \textbf{coherence} metrics for two datasets, which means \textit{THINK} tends to better understand the context and generate the coherent response. For the details of \textbf{avg(B)} (word level), \textbf{avg(E)} (word embedding level) and \textbf{coherence} (sentence embedding level), the \textit{THINK} also get a good performance in virtually all cases.

We also tested the impact of $head$ and $p$ on \textbf{distinct}-$n$), and found that $head$ increases the number of keywords that the model pays attention to and $p$ appear to improve the ability of understanding the keywords.

\subsubsection{Human evaluation analysis}
\label{sect:result-dia-human-eval}

\begin{table}[!h]
\renewcommand{\arraystretch}{1.3}
\centering
\begin{tabular}{lcccl}
\hline
\multirow{2}{*}{\textbf{Model}} & \multicolumn{3}{c}{\textbf{our model vs}(\%)} & \multirow{2}{*}{\textbf{kappa}} \\ \cline{2-4}
\ &\textbf{win} & \textbf{loss} & \textbf{tie} & \\
\hline
Seq2Seq     & 24.00 & 10.50 & 65.50 & 0.427 \\
CVAE        & 28.67 &  2.33 & 69.00 & 0.436 \\
TransFM     & 26.83 &  4.34 & 68.83 & 0.476 \\
CMHAM       & 26.17 &  5.83 & 68.00 & 0.455 \\
\hline
\hline
Seq2Seq     & 26.83 &  8.17 & 65.00 & 0.475 \\
CVAE        & 27.17 &  4.00 & 68.83 & 0.485 \\
TransFM     & 25.33 &  5.17 & 69.50 & 0.483 \\
CMHAM       & 23.83 &  7.33 & 68.83 & 0.491 \\
\hline
\end{tabular}
\caption{The human evaluation results on DailyDialog dataset (up) and CornellMovie dataset (down). }
\label{tab:human-evaluation-results}
\end{table}

We use human evaluation to further evaluate our model and baseline models.
The results of human evaluation are shown in Table \ref{tab:human-evaluation-results}.
In this table, the number under label \textbf{win} means that the percentage of sample responses generated by our model is better than that of the baseline model.
From Table \ref{tab:human-evaluation-results}, we can see that the percentage of the \textbf{win} is always larger than that of the \textbf{loss}, which demonstrates that our \textit{THINK} significantly outperforms baselines.
The \textbf{kappa} denotes the Kappa coefficient \cite{kappa-value}, used for consistency check of different annotators. The possible results of \textbf{kappa} are between -1 and 1, which demonstrates that the consistency degree is moderate.

\subsubsection{Dialog sample analysis}
\label{sect:result-dia-sample}

\begin{table}[!h]
\renewcommand{\arraystretch}{1.3}
\centering
\begin{tabular}{rp{5.5cm}}
\hline
Context:     & there are cockroaches in my room . \\
\hline
Seq2Seq:     & sorry, sir, sir, sir, sir, sir, sir, sir, sir, sir, sir, sir, sir  \\
CVAE:        & you sure, sir, but, 'm n't your but you have  \\
TransFM:     & there are many cities.  \\
CMHAM:       & why are you going to tell her that he is too late.  \\
             \hline
THINK:    & \textbf{what's the big deal ?}  \\
\hline
\hline
Context:     & are you a \textbf{goal-oriented} person ? \\
\hline
Seq2Seq:     & no, i don't. i don't know what i do.  \\
CVAE:        & , i am. that've make an collect. no ...  \\
TransFM:     & yes, i do. i was thinking of the whole day, yes.  \\
CMHAM:       & yes, it's all.  \\
\hline
THINK:    & yes, i am. that's why \textbf{i make plans before i do anything.}  \\
\hline
\end{tabular}
\caption{The responses produced by the proposed models and comparison models.}
\label{tab:dialogue-samples}
\end{table}

\begin{table*}[!t]
\renewcommand{\arraystretch}{1.3}
\centering
\small
\begin{tabular}{lccccccc}
\hline
\textbf{Model} &\textbf{q\_phrase-3} & \textbf{q\_phrase-4} & \textbf{q\_phrase-5} &  & \textbf{q\_phrase-3} & \textbf{q\_phrase-4} & \textbf{q\_phrase-5}\\
\hline
Seq2Seq    & 0.1554 & 0.1538 & 0.1170  & & 0.1083 & 0.1055 & 0.0770 \\
-w. \textit{TG} & \textbf{0.4069} & \textbf{0.4082}  & \textbf{0.3543} &  & \textbf{0.2730} & \textbf{0.2477} & \textbf{0.1623} \\
\hline
CVAE     & 0.2453 & 0.1513 & 0.0985 &  & 0.1766  & 0.0744  & 0.0234 \\
-w. \textit{TG}    & \textbf{0.2629} & \textbf{0.1979}  & \textbf{0.1191} &  & \textbf{0.1857} & \textbf{0.1474}  & \textbf{0.0753} \\
\hline
TransFM     & 0.2946 & 0.2349 & 0.1542  &  & 0.2533 & 0.1953 & 0.1043 \\
-w. \textit{TG} & \textbf{0.3709} & \textbf{0.3615}   & \textbf{0.3084} &  & \textbf{0.3179} & \textbf{0.2882} & \textbf{0.2192} \\
\hline
CMHAM    & 0.3053 & 0.2739 & 0.1996  &  & 0.2245 & 0.1946 & 0.1293 \\
-w. \textit{TG}   & \textbf{0.4066} & \textbf{0.4678}  & \textbf{0.4734}  &  & \textbf{0.3711} & \textbf{0.3893} & \textbf{0.3633} \\
\hline
\end{tabular}
\caption{Results of \textbf{q\_phrase}-$n$ on DailyDialog dataset (left) and CornellMovie dataset (right).}
\label{tab:ablation-TG}
\end{table*}
Table \ref{tab:dialogue-samples} shows several examples of conversations, which illustrates that the response generated by our model has a higher quality.
The first example demonstrates the ability of semantics-understanding of \textit{THINK}.
Observing the responses of each model, only ``\textit{sorry}" generated by Seq2Seq can have a relationship with context, but the generated text is extremely poor.
Our proposed \textit{THINK} can better understand the semantics of context through keywords and gives a more relevant and better response.
The second example shows the advantages of  \textit{THINK} over other models.
We can see that the \textit{THINK} model can extract core semantic features from context sentence.
Meanwhile, the \textit{TG} framework makes sure that the generated response will be non-repeated and consistent through the constrict of previous generated information.

\subsection{Ablation study}
\label{sect:classi}

\subsubsection{Teamwork generation framework}
We employ each baseline model as a generator in \textit{TG} framework to build a comparison.
The results of \textbf{q\_phrase}-$n$ are shown in Table \ref{tab:ablation-TG}, and the better results are marked through bold font. For instance, the first line in Table \ref{tab:ablation-TG} reports the \textbf{q\_phrase}-\{3,4,5\} of ``Seq2Seq'', and the second line shows the results of ``Seq2Seq'' with the \textit{TG} framework. From the data in the Table \ref{tab:ablation-TG}, we found that the baseline models get better performance on \textbf{q\_phrase}-$n$ while they equipping the \textit{TG} framework. These results demonstrate the effectiveness of the \textit{TG} framework in improving the \textbf{q\_phrase}-$n$ of generated responses.

Then, we extract some dialogue cases based on the \textbf{q\_phrase}-4 and \textbf{distinct}-4 from baseline models and baseline models with \textit{TG} framework, which shown in Table \ref{tab:ablation-TG-samples}. The generated responses we extracted are almost with high \textbf{distinct}-4 but low \textbf{q\_phrase}-4, which could explicitly show the strength of our \textit{TG} framework and prove the effectiveness of our proposed \textbf{q\_phrase}-$n$ metric in evaluating the grammatical quality.

\begin{table}
\renewcommand{\arraystretch}{1.3}

\centering
\begin{tabular}{rp{6cm}}
\hline
\textbf{model} & \textbf{generated response} \\\hline
Seq2Seq     & oh, i'm sorry. \underline{i'm sorry. i'm sorry.} \\
-w. \textit{TG} & i am sorry, there is no window seat.\\\hline
TransFM     & well, now \underline{we don't what} can we do? \\
-w. \textit{TG} & well, now we're stuck. what can we do?\\
\hline
\end{tabular}
\caption{The responses produced by the baseline models without \textit{TG} framework and with \textit{TG} framework.}
\label{tab:ablation-TG-samples}
\end{table}

\subsubsection{Semantics Extractor}
We adopt a topic classification task to illustrate that $DiscretelyDC$ is powerful in extracting semantic keywords, and evaluate the effectiveness of results based on precision, recall, F1-measure(F1) and accuracy.

\begin{figure}[ht]
\centering
\includegraphics[scale=0.7]{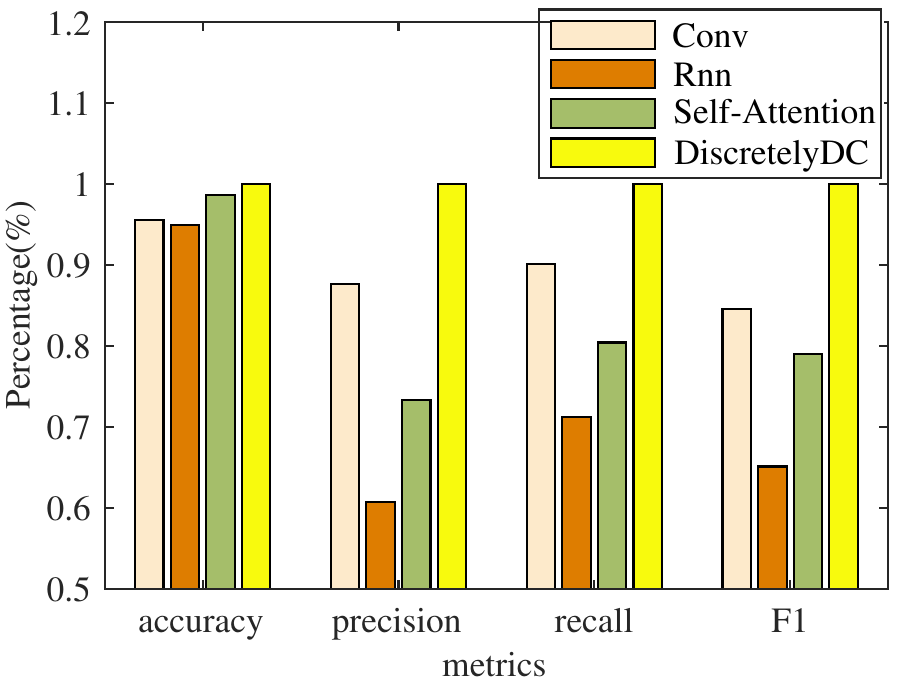}
\caption{Results of topic classification task}
\label{fig:ablation-SE-classification}
\end{figure}

To verify the advantages of $DiscretelyDC$, we use the results of $DiscretelyDC$ as the benchmark, and then calculate the percentage of results of other models to the benchmark.
As shown in Figure \ref{fig:ablation-SE-classification}, all models have a small gap on accuracy, and the self-attention (Transformer) and $DiscretelyDC$ results are similar, which is only a little better than the ordinary convolution (Conv) and Gate Recurrent Unit (Rnn).
However, considering the precision, recall, and f-values (F1) for each category, the result of $DiscretelyDC$ is the best.

\begin{figure}[ht]
\centering
\includegraphics[scale=0.85]{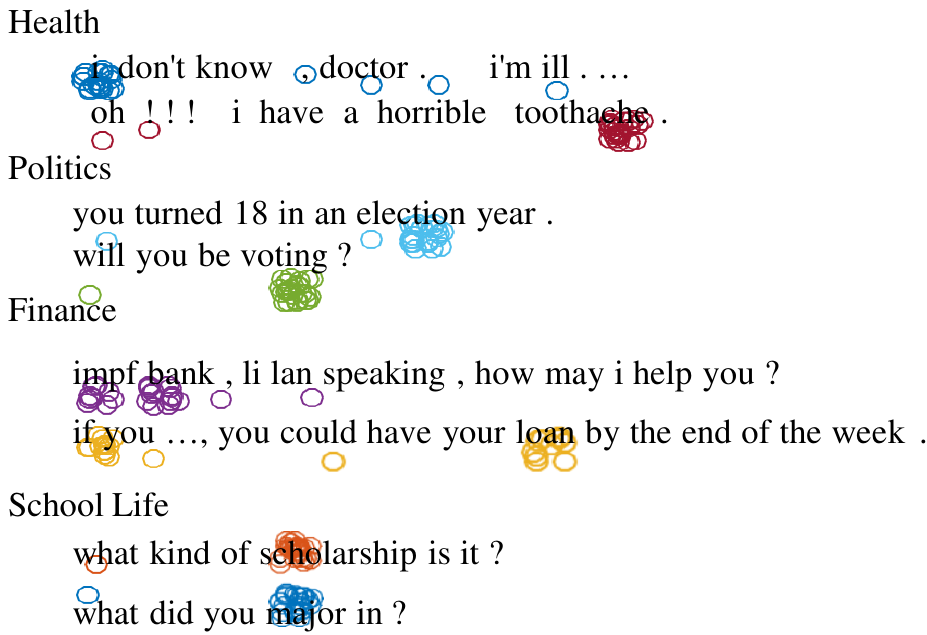}
\caption{Visualization of middle processing in $DiscretelyDC$}
\label{fig:ablation-SE-visualization}
\end{figure}

In order to observe the $DiscretelyDC$ in detail, we extract the middle process results of it. Then we extract a few clear samples and draw them as the following Figure \ref{fig:ablation-SE-visualization} after statistical process. 
Since the actual coordinate values are all integers, it is difficult to watch the concern degree of tokens. Therefore, we increase the abscissa of each point by random numbers within range -0.25-0.25, while the ordinate adds a random number between -0.125 and 0.125 to make the difference clearer. 
In Figure \ref{fig:ablation-SE-visualization}, `\textit{Health}', `\textit{Politics}' , `\textit{Finance}', `\textit{School Life}' represents the topic of sentences. We selected two sentences for each topic, and marked the keywords that the $DiscretelyDC$ focuses on. For example, we extracted two samples (``\textit{what did you major in ?}'' and ``\textit{what kind of schoolarship is it ?}'') for the `\textit{School Life}' topic. Figure \ref{fig:ablation-SE-visualization} shows that the $DiscretelyDC$ can focus on some keywords of related topics. For instance, the $DiscretelyDC$ finds `\textit{major}' and `\textit{scholarship}' for `\textit{School Life}' topic. Therefore, the \textit{SE} is able to directly extract semantic keywords-combination, which can replace the complex context vectors for providing semantic information in dialogue generation task.

\section{Related work}

\subsection{Dialogue generation methods}

Since the early dialogue models often generate general and dull responses, many works enrich the context vector with external information to address this problem in recent years. Taking a panoramic view of these works, there are three main directions for generating diverse responses: 1) using attention mechanism ~\cite{attention-origin-bahdanau15,attention-origin-luong15,hard-attention,transformer-base-17,cmham-multi-head-attention}, 2) using variational auto-encoder (VAE) structure ~\cite{vae-bowman16,cvae-zhao17,haofu-19,Guxiaodong-19} and 3) using external information ~\cite{external-info-jiwei,external-info-Marjan,external-info-Bernd}.
However, these methods complicate the context vector, which makes the semantics of context get deviation from original semantics of contexts, which leads to generate diverse but incoherent responses.

Therefore, two kinds of methods were proposed recently to address the diverse but incoherent responses problem. One introduced the reinforcement learning ~\cite{rl-1,RL-Seq2seqCo-Zhang2018,RL-P2BOT-Liu2020}, the other utilized the generative adversarial network ~\cite{gan-ljw,GAN-DialogWAE-Gu2019,GAN-PosteriorGAN-Feng2020}.
However, no matter how complex these approaches are, it always employs a simple decoder, which could not effectively utilize the previous generated information, and then often tends to generate the duplicate and self-contradict responses. Different from the existing dialogue models, we proposed \textit{TG} framework to utilize the previous generated information directly and explicitly during generating process.

\subsection{Convolution used in NLP}

Convolution neural network (CNN) was first introduced in sentence classification task \citep{TextCNN-kim-2014,DCNN-Kalchbrenner-2014,SemanticParsing-yih-2014,SemanticRepresentations-shen-2014}.
However, the main problem that uses the CNN in NLP task lies in CNN's  inability to capture long-distance dependent features. Nowadays this problem can be easily addressed by two groundbreaking methods: using hierarchical model structure ~\cite{ ARC-hu-nips-2014,gateCNN-Meng-2015,deconvAutoEncoder-Zhang-NIPS-2017} and using novel convolution kernel (\emph{e.g.} Dilated Convolution ~\cite{DilatedConv-2016-ICLR}, Depthwise Separable Convolutions ~\cite{DepthwiseConv-Kaiser-2018}). 
In addition, CNN could provide global information of the sentence, and allows parallelization on each element in the sentence, which is helpful for some NLP tasks \citep{Accelerate-DilatedConv-Kalchbrenner-2016,conv-seq2seq}.
These works enrich the theoretic foundation of CNN models and indicate that CNN models are applicable for many NLP tasks.

\section{Conclusion}

In this paper, we propose a dialogue model named \textit{THINK} (\textit{Teamwork generation Hover Impressiveness Noticeable Keywords}) to generate grammatically correct and coherent responses. The \textit{THINK} is based on the proposed \textit{Teamwork generation} framework and \textit{Semantics Extractor}. The \textit{TG} framework is proposed for addressing duplicate and self-contradict responses caused by simple decoder. The \textit{SE} is presented to overcome the incoherent responses caused by the complex context vectors. In conversation experiments, both automatic evaluation results and results of human evaluation are given to show the advantages of our \textit{THINK}. We did ablation study to illustrate that \textit{TG} framework effectively boosts the quality of responses and \textit{SE} makes it easy to understand semantics.

\bibliographystyle{acl_natbib}
\bibliography{anthology,acl2021}

\end{document}